\documentclass[letterpaper]{article} % DO NOT CHANGE THIS
\usepackage{aaai25}  % DO NOT CHANGE THIS
\usepackage{times}  % DO NOT CHANGE THIS
\usepackage{helvet}  % DO NOT CHANGE THIS
\usepackage{courier}  % DO NOT CHANGE THIS
\usepackage[hyphens]{url}  % DO NOT CHANGE THIS
\usepackage{graphicx} % DO NOT CHANGE THIS
\usepackage{natbib}  % DO NOT CHANGE THIS AND DO NOT ADD ANY OPTIONS TO IT
\usepackage{caption} % DO NOT CHANGE THIS AND DO NOT ADD ANY OPTIONS TO IT
\usepackage{algorithm}
\usepackage{algorithmic}

\usepackage[dvipsnames]{xcolor}
\definecolor{lightBlue}{RGB}{59,207,240}
\usepackage{newfloat}
\usepackage{listings}
\usepackage{color, multirow, multicol, verbatim, threeparttable, diagbox, makecell, booktabs, colortbl}

\DeclareCaptionStyle{ruled}{labelfont=normalfont,labelsep=colon,strut=off} % DO NOT CHANGE THIS
\floatstyle{ruled}
\newfloat{listing}{tb}{lst}{}
\floatname{listing}{Listing}
\title{ODDN: Addressing Unpaired Data Challenges in Open-World Deepfake Detection on Online Social Networks}
\author{
    Renshuai Tao\textsuperscript{\rm 1}, Manyi Le\textsuperscript{\rm 1}, Chuangchuang Tan\textsuperscript{\rm 1}, Huan Liu\textsuperscript{\rm 1}, Haotong Qin\textsuperscript{\rm 2}, Yao Zhao\textsuperscript{\rm 1}
}
\affiliations{
    \textsuperscript{\rm 1}Beijing Jiaotong University\\
    \textsuperscript{\rm 2}ETH Zürich, Switzerland\\
    \emph{rstao@bjtu.edu.cn}
}

\usepackage{bibentry}
\begin{document}

\maketitle

\begin{abstract}
Despite significant advances in deepfake detection, handling varying image quality, especially due to different compressions on online social networks (OSNs), remains challenging. Current methods succeed by leveraging correlations between paired images, whether raw or compressed. However, in open-world scenarios, paired data is scarce, with compressed images readily available but corresponding raw versions difficult to obtain. This imbalance, where unpaired data vastly outnumbers paired data, often leads to reduced detection performance, as existing methods struggle without corresponding raw images. To overcome this issue, we propose a novel approach named the open-world deepfake detection network (ODDN), which comprises two core modules: open-world data aggregation (ODA) and compression-discard gradient correction (CGC). ODA effectively aggregates correlations between compressed and raw samples through both fine-grained and coarse-grained analyses for paired and unpaired data, respectively. CGC incorporates a compression-discard gradient correction to further enhance performance across diverse compression methods in OSN. This technique optimizes the training gradient to ensure the model remains insensitive to compression variations. Extensive experiments conducted on 17 popular deepfake datasets demonstrate the superiority of the ODDN over SOTA baselines.
% The code is available at \url{https://anonymous.4open.science/r/ODDN/}.

\end{abstract}

\section{Introduction}

With the rapid development of deep learning-based generation technology\cite{zhou2023generative,yu2023reinforcement,zhang2023encrypted,zhang2024dibad,zhao2023istvt}, AI-generated images are increasingly appearing on social media platforms like Twitter and WeChat. While these images enhance creativity and enjoyment, they also introduce significant safety risks. The ability to create highly realistic images easily has raised concerns about misinformation, privacy, and security. Deepfakes can be used to spread false information, creating fake news that misleads the public. Additionally, these images can be exploited for malicious purposes such as identity theft, fraud, and cyberbullying, amplifying the potential for harm across social media platforms\cite{asnani2023reverse,vice2024bagm,ding2023tmg,zeng2024towards}.

\begin{figure}[!t]
\vspace{0.2in}
  \centering
   \includegraphics[scale=0.42]{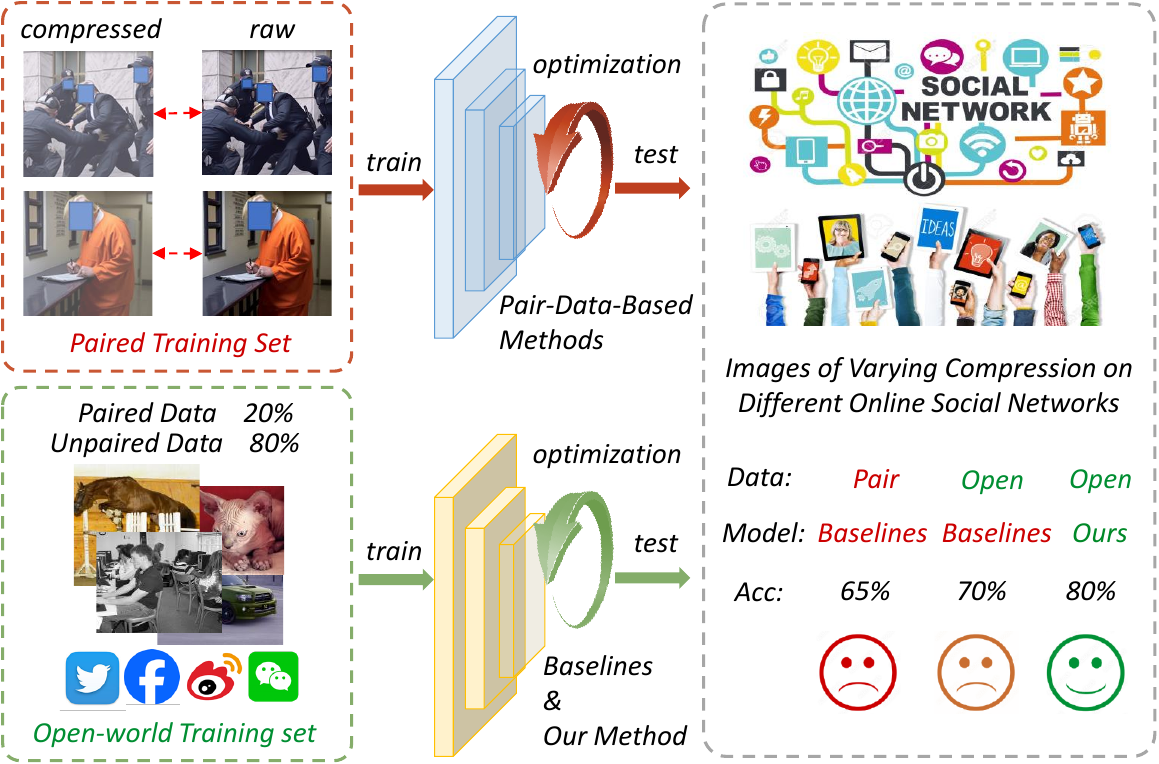}
   \caption{Comparison of training data and models between traditional and open-world scenarios. Open-world scenarios require models to effectively manage the unpaired data sources, addressing the inherent variability and complexity.}
   \label{fig:fig1}
  % \vspace{-0.45 cm}
  \vspace{-0.18in}
\end{figure}

Despite significant progress in detecting AI-generated images, detecting forged images on online social networks (OSN) has received relatively little attention. Images on these platforms are often subjected to various compression methods \cite{dzanic2020fourier,wu2023robust}, complicating detection efforts. Compression techniques employed by platforms like Twitter and WeChat degrade image quality and obscure manipulation indicators, making forgery identification more challenging. As a result, developing robust detection methods capable of overcoming the unique challenges posed by social media compression remains an urgent research focus in image forensics.

Current methods \cite{wu2022robust,le2023quality,liu2023generating,le2024gradient} for detecting forged images in compressed formats typically rely on paired data (compressed images and their corresponding originals) for training. These approaches focus on identifying feature correlations between paired data, achieving notable performance improvements under specific compression methods. However, in real-world OSN scenarios, obtaining the original images corresponding to compressed ones is often impractical, leading to a significant imbalance between paired and unpaired data, with unpaired data far outnumbering paired data. The abundance of unpaired data, whether real or fake, compressed or raw, contains valuable evidence that can aid in authenticity differentiation. However, methods reliant on paired data often fail to effectively integrate this unpaired data, resulting in the loss of critical information. Additionally, existing strategies that focus on fine-grained data associations between compressed images and their corresponding raw versions struggle to address coarse-grained connections among unpaired data. Therefore, exploring robust deepfake detection techniques that can handle open-world scenarios with imbalanced data is crucial for controlling the spread of forged information on various OSN platforms.

In this paper, we first identify the challenge of open-world deepfake detection on OSN, where available training data are often unpaired. Traditional pair-data-based methods are unsuitable for this scenario because compressed images typically lack the corresponding original images, making it difficult to apply these traditional approaches effectively. To address this imbalanced dilemma, we propose a novel method called the open-world deepfake detection network (ODDN), which comprises two core modules: open-world data aggregation (ODA) and compression-discard gradient correction (CGC). The ODA module tackles the challenge of aligning true and false sample features across different types of data, while the CGC module addresses the issue of poor gradient optimization direction when removing compression-sensitive information during training.

Specifically, the ODA module handles paired and unpaired input data from open-world OSN with distinct processing approaches. For a small quantity of paired data (20\%), the ODA module exploits fine-grained correlations between the compressed images and their corresponding originals. For the remaining 80\% unpaired data, it establishes coarse-grained correlations by clustering the true and false images. Meanwhile, the CGC module ensures the model's insensitivity to compression, which is crucial for effectively handling various compression methods in open-world OSN scenarios. It adopts PCGrad to align and facilitate interactions between distinct gradients, ensuring that the optimization process remains focused on directions that positively impact the main task of the real/fake discrimination.

To comprehensively evaluate the effectiveness of the proposed ODDN, we designed an innovative training data setup to simulate an open-world OSN environment, where unpaired data (80\%) far outnumber paired data (20\%). Specifically, we compressed a small portion of the training data, typically used for forgery detection tasks, to create paired data, which was then combined with the remaining unpaired data to form the training set. We trained all baseline models and our method on this same training set and assessed their performance under two different test conditions: one aligned with the compression level of the training set and the other unrelated. Our evaluation involved 17 popular GAN-based datasets across these two test settings. The final results demonstrate that our model significantly outperforms existing state-of-the-art models, showcasing its superior effectiveness in OSN. Our contributions are summarized blow:
Here’s a reconstructed version of the three contributions:
\begin{itemize}
\item We introduce the challenge of unpaired data in deepfake detection within open-world scenarios on OSN by designing a novel setup that simulates these environments, offering a valuable benchmark for future research.
\item To handle this dilemma, we propose the ODDN, comprising ODA for optimizing artifact feature alignment in unpaired data scenarios, and CGC for reducing gradient biases when removing compression-related information, thereby enhancing detection robustness and adaptability.
\item Comprehensive experiments have validated the effectiveness of ODDN across 17 popular datasets under various test settings, demonstrating superior performance in detecting deepfakes on OSN compared to SOTA baselines.
\end{itemize}

\begin{figure*}[!t]
\vspace{-0.5in}
\centering
\includegraphics[width=\textwidth]{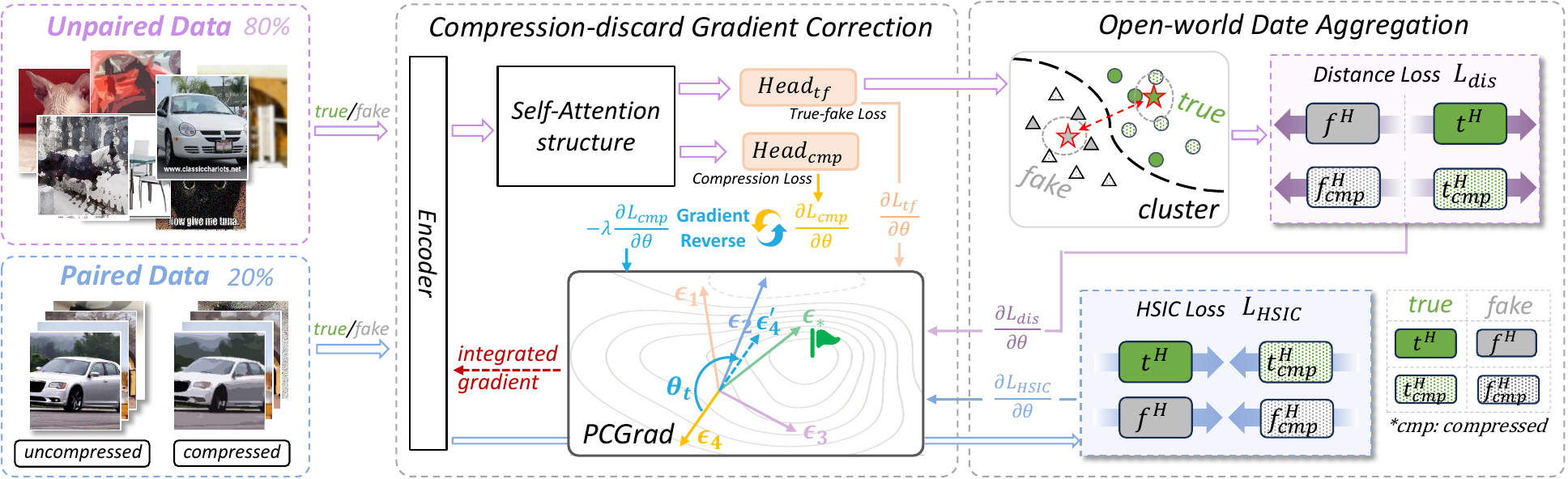} % Reduce the figure size so that it is slightly narrower than the column.
\caption{Overview of the proposed Open-world Deepfake Detection Network (ODDN). The ODDN contains two core modules: Open-world Data Aggregation (ODA) and compression-discard Gradient Correction (CGC). The ODA module focuses on aligning unpaired data in open-world OSN by leveraging the feature-center points. The CGC module ensures the proper optimization direction through integrated gradient correction during the process of removing compression-sensitive information.}
\vspace{-0.18in}
\label{fig2}
\end{figure*}

\section{Related Work}
Here, we delve into a detailed survey of compressed deepfake detection: AI-Generated image detection and OSN.

\subsection{AI-Generated Image Detection}
Various strategies have been employed to enhance the generalization of detectors to unseen sources. These strategies include diversifying training data through augmentation methods \cite{wang2020cnn,wang2021representative}, adversarial training \cite{chen2022self}, reconstruction techniques \cite{cao2022end,he2021beyond}, fingerprint generators \cite{jeong2022fingerprintnet}, and blending images \cite{shiohara2022detecting}. Specific methodologies such as BiHPF \cite{jeong2022bihpf} amplify artifacts' magnitudes through two high-pass filters. FreGAN \cite{jeong2022frepgan} addresses the overfitting of training sources by mitigating the impact of frequency-level artifacts through frequency-level perturbation maps.
Ju et al.\cite{ju2022fusing} integrate global spatial information and local informative features in a two-branch model. AltFreezing by Wang et al.\cite{wang2023altfreezing} leverages both spatial and temporal artifacts for Face Forgery Detection.
Approaches by Ojha et al.\cite{ojha2023towards} and Tan et al.\cite{Tan2023CVPR} utilize feature maps and gradients, respectively, as general representations. DIRE by Wang et al.~\cite{Wang_2023_ICCV} introduces a novel image representation by measuring the feature distance between an input image and its reconstruction counterpart, aiming to alleviate generalization issues.

\subsection{Online Social Networks (OSN)}
The widespread use of various online social network platforms, such as Facebook, WhatsApp, WeChat, and Weibo, has significantly facilitated the dissemination and sharing of images. However, numerous studies\cite{sun2016processing,sun2020robust} have indicated that almost all OSN manipulate uploaded images in a lossy manner, introducing noise that can severely impact the effectiveness of forensic methods. For instance, as identified in the paper\cite{sun2021optimal}, Facebook's image manipulation process comprises four main stages: format conversion, resizing, enhancement filtering, and JPEG compression. Initially, the uploaded image is converted into the pixel domain, with truncation ensuring pixel values remain within the [0, 255] range. If the image resolution exceeds 2048 pixels, resizing is applied. Subsequently, selected blocks within the image undergo adaptive and complex enhancement filtering, the specifics of which are challenging to ascertain due to their adaptiveness. Finally, the image is subjected to JPEG compression, with the quality factor (QF) being adaptively determined based on the image content. While image manipulations vary across different OSN platforms, mainstream OSN exhibit many similarities in their operations, such as the ubiquitous use of JPEG compression. Therefore, it is crucial to study the robustness of the deepfake detection models under different JPEG compressions on various online social network platforms.

\section{Method}
In this section, we elaborate on the proposed Open-world Deepfake Detection Network (ODDN) previously introduced. First, we formalize the task of open-world deepfake detection on OSN. Subsequently, we provide a comprehensive overview of the ODDN framework. Finally, we introduce two core modules: Isolated Data Aggregation (IDA) and compression-discard Gradient Correction (CGC).

\subsection{Problem Definition}
Due to the paucity of paired data, specifically an original resolution Deepfake image and its compressed version, the focus of our work is distinct from previous studies. Consequently, to concisely and vividly explain our method, we must adapt the common problem definition used in previous works. Given a dataset $\mathcal{D} = \{(x_i, y_i)\}_{i=1}^{N}$, it comprises two types of images: real images $x^r$ and Deepfake images $x^f$, with the corresponding labels $y \in \{0, 1\}$ representing real or fake. Subsequently, 20\% of the data in $\mathcal{D}$ is randomly chosen to undergo a JPEG compression operation, denoted as $P_c$, maintaining about 60\% image quality. The original versions of these images and the rest of the data in $\mathcal{D}$ are denoted as $P$ and $\tilde{P}$, respectively. Therefore, we can rewrite the composition of the dataset as $\mathcal{D}_{train} = P\cup P_c \cup \tilde{P}$. 

As for the inference stage, there are two types: quality-aware and quality-agnostic inference. In the first type, quality-aware inference, the images of the testing set $\mathcal{D}_{test}$ are compressed using the same operation as $P_c$. In the second type, quality-agnostic inference, the images of the testing set $\mathcal{D}_{test}$ are compressed using various operations to mimic an open-world scenario where the compression type is unknown.
The lack of paired data makes it much harder to improve model robustness, creating significant challenges for various efforts, including state-of-the-art approaches.

\subsection{Network Framework}
As illustrated in Figure \ref{fig2}, the proposed ODDN framework comprises two core modules: ODA and CGC. Within the ODA module, data operations are conducted on two types of datasets: 80\% unpaired data and other 20\% paired data. The unpaired data is aligned using feature-center points, while the paired data is processed using a classic method. The CGC module ensures optimization in the correct direction by integrating gradient correction during the removal of compression-sensitive information. By leveraging a multi-layer adversarial learning mechanism, the framework effectively confounds compression-related characteristics, allowing the detection model to focus on compression-insensitive information. And self-Attention structure is employed to attend to distinct features that are required by different down-stream tasks. This approach significantly enhances the generalization capability of deepfake detection models.

\subsection{Open-world Data Aggregation}
The ODA module primarily involves distinct processing of unpaired and paired input data in open-world scenarios.

\textbf{Solution for the unpaired data.} Unlike paired data, unpaired data lacks fine-grained correlations between the compressed and corresponding original images, requiring alternative alignment methods to effectively utilize the abundant unpaired data resources.
We aim to establish coarse-grained correlations in unpaired data, which lack strong connections, by clustering true and false images. Specifically, for a given batch of input, we calculate four aggregation centers: the real images $C^{t}$, the compressed real images $C^{t}_{cmp}$, the fake images $C^{f}$, and the compressed fake image $C^{f}_{cmp}$. The definition formulas for these four quantities are as follows:
\begin{equation}
    C^{t} = \frac{\Sigma^{N^t}_{i=1}h^{H}_{i}}{N^t},\ \ \    
    C^{t}_{cmp} = \frac{\Sigma^{N^t_{cmp}}_{i=1}h^{H}_{i}}{N^t_{cmp}}
\end{equation}

\begin{equation}
    C^{f} = \frac{\Sigma^{N^f}_{i=1}h^{H}_{i}}{N^f},\ \ \ 
    C^{f}_{cmp} = \frac{\Sigma^{N^f_{cmp}}_{i=1}h^{H}_{i}}{N^f_{cmp}}
\end{equation}
where $N^t$, $N^t_{cmp}$, $N^f$, $N^f_{cmp}$ represent the respective counts of images belonging to four distinct classes within a batch, $h_i^{H}$ is the feature obtained from the self-attention block.

Subsequently, we enlarge the separation among these cluster centers to improve the distinction between real and fake images, making it easier for the detection model to accurately recognize them. Specifically, to enable our model to effectively classify deepfakes, even in their compressed forms, we strategically increase the distance between the cluster centers of \(C^{t}\) and \(C^{f}\), as well as the distance \(S_{cmp}\) between the compressed clusters \(C^{t}_{cmp}\) and \(C^{f}_{cmp}\). The detailed formulas are as follows:
\begin{equation}
    S = \frac{1}{1 + \Sigma^d_{i = 1}\sqrt{(C^{t}_i - C^{f}_i)^2}} 
\end{equation}

\begin{equation}
    S_{cmp} = \frac{1}{1 + \Sigma^d_{i = 1}\sqrt{(C^{t}_{i,cmp} - C^{f}_{i,cmp})^2}}
\end{equation}
where $d$ denotes the dimension of the hidden features.

The sum of these two types of distances, $S_{cmp}$ and $S$, can be considered as the alignment loss $\mathcal{L}_{dis}$ for unpaired data. This loss function not only increases the separation between real and fake images but also promotes the aggregation of images within the same class. For greater clarity, the alignment loss of unpaired data can be expressed as follows:
\begin{equation}
    \mathcal{L}_{unpair} = S + S_{cmp}
\end{equation}

\textbf{Solution for the paired data.} Following the previous work \cite{le2023quality}, we observe that the Hilbert-Schmidt Independence Criterion (HSIC), a metric for measuring correlation, is an effective method for maximizing dependency among images of varying quality. This approach allows the model to learn intricate distribution relationships between paired images. Given the scarcity and value of paired data, we continue to apply HSIC to paired data, as it is a method that can fully exploit the useful information within these pairs. The formula is as follows:
\begin{equation}
    \mathcal{L}_{pair} = \widehat{HSIC}(h^{E}_{c},\ h^{E})
\end{equation}
where $h^{E}_{c}$ and $h^{E}$ represent the features of the compressed image and the corresponding original image within the paired data, respectively, as output by the image encoder.

\begin{table*}[!t]
\vspace{-0.5in}
\Huge
    \centering
\setlength{\tabcolsep}{1.5pt}
\resizebox{\textwidth}{18mm}{
    \begin{tabular}{ c c c c c c c c c c c c c c c c c c  >{\columncolor{lightgray}}c}
     \bottomrule\hline
\multirow{2}*{Method}  & 
\multirow{2}*{ \makecell[c]{ InfoMax- \\  GAN} }  & 
\multirow{2}*{ \makecell[c]{ BE- \\  GAN} }  & 
\multirow{2}*{ \makecell[c]{ Cramer- \\ GAN} }  & 
\multirow{2}*{ \makecell[c]{ Att- \\ GAN } } & 
\multirow{2}*{ \makecell[c]{ MMD- \\  GAN} }  & 
\multirow{2}*{ \makecell[c]{ Rel- \\  GAN} } & 
\multirow{2}*{ \makecell[c]{ S3- \\  GAN} }  & 
\multirow{2}*{ \makecell[c]{ SNG- \\  GAN} }  & 
\multirow{2}*{ \makecell[c]{ STG- \\  GAN}  } & 
\multirow{2}*{ \makecell[c]{ Pro- \\  GAN} } &
\multirow{2}*{ \makecell[c]{ Style- \\  GAN} }&
\multirow{2}*{ \makecell[c]{ Style- \\  GAN2} } &
\multirow{2}*{ \makecell[c]{ Big- \\  GAN} } &
\multirow{2}*{ \makecell[c]{ Cycle- \\  GAN} } &
\multirow{2}*{ \makecell[c]{ Star- \\  GAN} } &
\multirow{2}*{ \makecell[c]{ Gau- \\  GAN} } &
\multirow{2}*{ \makecell[c]{ Deep- \\  fake} } &
Mean \\ &&&&&&&&&&&&&&&&&& Acc \\
\bottomrule\hline
MesoNet\shortcite{afchar2018mesonet}           & 50.5 & 50.6 & 50.0 & 50.5 & 50.2 & 50.4 & 49.3 & 50.2 & 50.9 & 51.4 & 51.5 & 54.2 & 52.0 & 53.4 & 50.6 & 53.4 & 50.0 & 51.2\\
FF++  \shortcite{rossler2019faceforensics++}   & 74.4 & 30.6 & 75.5 & 64.2 & 76.3 & 61.5 & 54.9 & 71.8 & 82.2 & 90.4 & 60.4 & 65.2 & 60.5 & 80.0 & 74.4 & 72.3 & 51.0 & 67.3\\
F3Net \shortcite{qian2020thinking}            & 65.9 & 42.6 & 68.9 & 55.9 & 63.7 & 56.3 & 53.1 & 62.1 & 74.9 & 84.1 & 56.7 & 60.4 & 56.1 & 77.8 & 71.2 & 68.6 & 50.4 & 63.2\\
MAT  \shortcite{zhao2021multi}                & 54.5 & 49.8 & 59.7 & 50.1 & 57.8 & 50.8 & 52.8 & 52.8 & 56.7 & 85.7 & 52.4 & 53.1 & 52.9 & 72.2 & 57.6 & 67.6 & 51.1 & 57.7\\
SBIs \shortcite{shiohara2022detecting}        & 56.6 & 51.9 & 63.4 & 50.1 & 59.3 & 50.6 & 62.2 & 52.1 & 53.0 & 88.4 & 51.2 & 52.4 & 55.4 & 74.8 & 53.6 & 78.3 & 51.1 & 59.3\\
ADD  \shortcite{woo2022add}                  & 52.0 & 51.0 & 59.0 & 50.7 & 57.2 & 52.7 & 44.7 & 52.3 & 53.1 & 70.9 & 48.0 & 48.4 & 51.7 & 72.4 & 55.7 & 64.7 & 51.3 & 55.2\\
QAD  \shortcite{le2023quality}               & 74.8 & 53.7 & 79.6 & 60.1 & 78.3 & 66.5 & 56.0 & 76.3 & 80.4 & 86.3 & 55.4 & 57.2 & 59.1 & 77.1 & 79.9 & 65.8 & 55.8 & 69.2\\
ODDN (\textbf{ours})                      & 73.1 & 42.3 & 76.1 & 71.2 & 75.9 & 72.5 & 60.5 & 75.5 & 85.0  & 91.3 & 64.5 & 69.4 & 64.3 & 80.8 & 78.0 & 77.3 & 54.3 & \textbf{71.4}\\
\bottomrule
    \end{tabular}
}
  \caption{The \textbf{quality-aware} experimental results across 17 well-known datasets under the \textbf{2-class training data setting}. The gray column on the far right of all tables represents the average accuracy performance of the models across 17 diverse datasets. This column is not merely a summary of results but serves as a critical metric for assessing the overall generalization ability.}
  \label{2class Quality-aware}
  \vspace{-0.05in}
\end{table*}
\begin{table*}[!t]
\Huge
    \centering
\setlength{\tabcolsep}{1.5pt}
\resizebox{\textwidth}{18mm}{
%\resizebox{2.1\columnwidth}{!}{
    \begin{tabular}{ c c c c c c c c c c c c c c c c c c  >{\columncolor{lightgray}}c}
     \bottomrule\hline
      %\multirow{3}*{Method} &\multicolumn{18}{c}{ Test Models}\\ 
\multirow{2}*{Method}  & 
\multirow{2}*{ \makecell[c]{ InfoMax- \\  GAN} }  & 
\multirow{2}*{ \makecell[c]{ BE- \\  GAN} }  & 
\multirow{2}*{ \makecell[c]{ Cramer- \\ GAN} }  & 
\multirow{2}*{ \makecell[c]{ Att- \\ GAN} } & 
\multirow{2}*{ \makecell[c]{ MMD- \\  GAN} }  & 
\multirow{2}*{ \makecell[c]{ Rel- \\  GAN} } & 
\multirow{2}*{ \makecell[c]{ S3- \\  GAN} }  & 
\multirow{2}*{ \makecell[c]{ SNG- \\  GAN} }  & 
\multirow{2}*{ \makecell[c]{ STG- \\  GAN } } & 
\multirow{2}*{ \makecell[c]{ Pro- \\  GAN} } &
\multirow{2}*{ \makecell[c]{ Style- \\  GAN} }&
\multirow{2}*{ \makecell[c]{ Style- \\  GAN2} } &
\multirow{2}*{ \makecell[c]{ Big- \\  GAN} } &
\multirow{2}*{ \makecell[c]{ Cycle- \\  GAN} } &
\multirow{2}*{ \makecell[c]{ Star- \\  GAN} } &
\multirow{2}*{ \makecell[c]{ Gau- \\  GAN} } &
\multirow{2}*{ \makecell[c]{ Deep- \\  fake} } &
Mean \\ &&&&&&&&&&&&&&&&&& Acc \\
\bottomrule\hline
MesoNet\shortcite{afchar2018mesonet}    & 49.5 & 46.2 & 52.6 & 51.3 & 53.0 & 53.8 & 50.4 & 51.8 & 54.2 & 53.3 & 49.6 & 53.9 & 55.1 & 50.9 & 52.3 & 51.7 & 45.0 & 51.4\\
FF++\shortcite{rossler2019faceforensics++}      & 69.5 & 26.9 & 80.3 & 66.8 & 79.2 & 69.9 & 56.2 & 75.1 & 84.4 & 93.6 & 62.5 & 60.8 & 58.5 & 80.9 & 78.5 & 71.00 & 52.8 & 68.7\\
F3Net\shortcite{qian2020thinking}      & 61.0 & 41.9 & 65.8 & 52.9 & 63.8 & 55.5 & 53.8 & 59.6 & 71.5 & 92.2 & 76.0 & 59.1 & 55.9 & 57.9 & 71.8 & 66.0 & 52.1 & 62.4\\
MAT\shortcite{zhao2021multi}       & 57.9 & 46.9 & 64.2 & 50.8 & 63.4 & 52.4 & 52.1 & 56.2 & 61.8 & 90.8 & 54.2 & 53.9 & 52.4 & 73.1 & 61.4 & 64.8 & 51.2 & 59.5\\
SBIs \shortcite{shiohara2022detecting}  & 60.2 & 55.7 & 74.4 & 50.2 & 67.1 & 54.6 & 61.4 & 53.0 & 57.2 & 96.0 & 57.4 & 53.0 & 55.4 & 77.6 & 60.1 & 74.9 & 50.6 & 62.5\\
ADD \shortcite{woo2022add}   & 51.7 & 50.7 & 57.3 & 51.3 & 55.9 & 52.4 & 45.2 & 51.2 & 52.4 & 73.5 & 49.9 & 50.1 & 52.2 & 70.7 & 54.4 & 66.4 & 51.2 & 55.3\\
QAD \shortcite{le2023quality}   & 79.9 & 37.5 & 79.5 & 67.4 & 76.8 & 71.7 & 58.0 & 79.0 & 83.5 & 92.7 & 64.7 & 68.7 & 64.0 & 81.8 & 80.3 & 66.3 & 52.9 & 70.9\\
ODDN (\textbf{ours})                      & 80.6 & 38.6 & 80.7 & 65.8 & 78.8 & 71.1 & 60.5 & 76.7 & 85.8  & 94.0 & 67.7 & 69.9 & 66.7 & 84.9 & 80.5 & 75.2 & 54.2 & \textbf{72.6}\\
\bottomrule
    \end{tabular}
}
  \caption{The \textbf{quality-aware} experimental results across 17 well-known datasets under the \textbf{4-class training data setting}.}
  \label{4class Quality-aware}
  \vspace{-0.18in}
\end{table*}

\subsection{Compression-discard Gradient Correction}
This module classifies true and false images using binary cross-entropy loss to distinguish deepfakes, thereby enhancing its ability to detect and identify synthetic content. This process can be formulated as follows:

\begin{equation}
    \mathcal{L}_{tf} = \mathcal{L}_{bce}(H_{tf}(h^{H}_{i}),\ y_i)
\end{equation}
where \(\mathcal{L}_{bce}\) represents the binary cross-entropy loss, \(H_{tf}\) is the head layers of the true/false classification, \(h^{H}_{i}\) is the feature obtained from the self-attention block, and \(y_i\) is the label for the corresponding input sample.

To effectively handle various compression methods in open-world OSN scenarios, the ideal criterion for discrimination should be insensitivity to compression. Thus, we exploit the adversarial learning mechanism, performing effective confusion for discarding the compression information. We assume that compressed images inherently carry a unique signature or fingerprint characteristic of the compression method used, such as JPEG compression. When training on a dataset of compressed images, the model may learn this fingerprint, potentially introducing biases and distorting performance. Our goal is to develop an encoder that maximizes the extraction of features related to fake artifacts while minimizing the inclusion of compression fingerprints. This approach enables the encoder to distinguish between real and fake images, regardless of the compression applied. Inspired by domain-adversarial training of neural networks, we introduce an additional downstream task and use a gradient reversal layer to achieve this goal.

Similarly, the compression-discard loss functions much like the true/false classification branch but differs in data processing. After passing through the image encoder, only the features of paired data are input the compression-discard branch, where they are assessed to determine whether they have been compressed. This operation is defined as follows:
\begin{equation}
    \mathcal{L}_{cmp} = \mathcal{L}_{bce}(H_{cmp}(h^{H}_{i}),\ y_i)
\end{equation}
where $y\in\{0, 1\}$, representing compressed or not and $H_{cmp}$ is the head of the compression-discard branch.

Furthermore, the gradient reversal layer inverts the gradient as it passes through. Consequently, when the gradient of \(\mathcal{L}_{cmp}\) propagates through the network, the gradients in the encoder and the compression-discard branch have opposite directions but the same magnitude. This operation forces the encoder to discard compression-related information, while the remaining components of the compression classification branch continue to be optimized for detection. The final loss function for the training process is a weighted sum of the above loss functions:
\begin{equation}
    \mathcal{L}_{all} = \mathcal{L}_{unpair} + \alpha\mathcal{L}_{pair} + \mathcal{L}_{tf} + \mathcal{L}_{cmp}
\end{equation}
where \(\alpha\) is hyper-parameter that balance the contributions of each component to the overall loss. It is worth noting that GCM effectively leverages valuable information, particularly by utilizing the numerous unpaired data. Additionally, the structure of the branches within GCM is flexible, allowing for the incorporation of other desired models.

However, \textbf{with many directions negatively correlated with the gradient direction of the loss \(\mathcal{L}_{cmp}\), how can we identify the most suitable direction}? Despite the aforementioned mechanism forcing the encoder to optimize in the reverse gradient direction of \(\mathcal{L}_{cmp}\), conflicts often arise between this direction and other gradients. Therefore, it's essential to find a way to align the reverse gradient with other gradients. PCGrad\cite{yu2020gradient} offers a solution by projecting conflicting gradients onto the normal vector of another, ensuring constructive interactions among non-conflicting gradients. Inspired by this, we exploit the conflicting gradients projection mechanism to align and facilitate interactions between distinct gradients, ensuring the optimization process remains focused on directions that positively impact the main task. The comprehensive gradient calculation formula is as follows:
\begin{equation}
    \nabla_E = \textbf{P}(\nabla(\mathcal{L}_{pair} + \mathcal{L}_{unpair} + \mathcal{L}_{tf}), -\nabla\mathcal{L}_{cmp})
\end{equation}
where \(\nabla_E\) represents the total gradient calculated for the encoder, ensuring that the gradients are optimized for optimal performance. The symbol \(\textbf{P}\) denotes the conflicting gradients projection, which is responsible for projecting conflicting gradients onto the normal vector of each other, thereby facilitating interactions among the gradients involved.

\begin{table*}[!ht]
\vspace{-0.5in}
\Huge
    \centering
\setlength{\tabcolsep}{1.5pt}
\resizebox{\textwidth}{18mm}{
%\resizebox{2.1\columnwidth}{!}{
    \begin{tabular}{ c c c c c c c c c c c c c c c c c c  >{\columncolor{lightgray}}c}
     \bottomrule\hline
      %\multirow{3}*{Method} &\multicolumn{18}{c}{ Test Models}\\ 
\multirow{2}*{Method}  & 
\multirow{2}*{ \makecell[c]{ InfoMax- \\  GAN } }  & 
\multirow{2}*{ \makecell[c]{ BE- \\  GAN} }  & 
\multirow{2}*{ \makecell[c]{ Cramer- \\ GAN} }  & 
\multirow{2}*{ \makecell[c]{ Att- \\ GAN } } & 
\multirow{2}*{ \makecell[c]{ MMD- \\  GAN} }  & 
\multirow{2}*{ \makecell[c]{ Rel- \\  GAN} } & 
\multirow{2}*{ \makecell[c]{ S3- \\  GAN} }  & 
\multirow{2}*{ \makecell[c]{ SNG- \\  GAN} }  & 
\multirow{2}*{ \makecell[c]{ STG- \\  GAN } } & 
\multirow{2}*{ \makecell[c]{ Pro- \\  GAN} } &
\multirow{2}*{ \makecell[c]{ Style- \\  GAN} }&
\multirow{2}*{ \makecell[c]{ Style- \\  GAN2} } &
\multirow{2}*{ \makecell[c]{ Big- \\  GAN} } &
\multirow{2}*{ \makecell[c]{ Cycle- \\  GAN} } &
\multirow{2}*{ \makecell[c]{ Star- \\  GAN} } &
\multirow{2}*{ \makecell[c]{ Gau- \\  GAN} } &
\multirow{2}*{ \makecell[c]{ Deep- \\  fake} } &
Mean \\ &&&&&&&&&&&&&&&&&& Acc \\
\bottomrule\hline
MesoNet\shortcite{afchar2018mesonet}     & 46.3 & 44.3 & 59.7 & 60.0 & 59.8 & 58.7 & 47.8 & 56.3 & 69.1 & 55.0 & 51.0 & 49.9 & 53.9 & 60.5 & 64.8 & 49.9 & 51.1 & 53.3\\
FF++\shortcite{rossler2019faceforensics++}        & 66.9 & 37.7 & 79.4 & 56.6 & 77.1 & 60.5 & 55.0 & 69.2 & 79.8 & 87.8 & 55.1 & 59.8 & 57.1 & 79.9 & 75.6 & 71.6 & 52.0 & 66.0\\
F3Net\shortcite{qian2020thinking}      & 58.0 & 48.8 & 61.9 & 51.5 & 59.3 & 53.2 & 52.0 & 54.9 & 61.1 & 83.4 & 52.7 & 54.9 & 55.0 & 73.7 & 65.9 & 66.7 & 52.4 & 59.3\\
MAT\shortcite{zhao2021multi}      & 54.2 & 49.9 & 59.6 & 50.5 & 57.6 & 51.2 & 52.1 & 52.7 & 57.8 & 86.1 & 52.3 & 53.0 & 52.7 & 70.3 & 58.2 & 68.0 & 51.3 & 57.7\\
SBIs \shortcite{shiohara2022detecting}   & 56.6 & 51.9 & 63.4 & 50.1& 59.3 & 50.6 & 62.2 & 52.1 & 53.0 & 88.6 & 51.3 & 52.4 & 55.7 & 76.0 & 53.9 & 78.1 & 51.2 & 59.4\\
ADD \shortcite{woo2022add}    & 51.8 & 50.9 & 59.0 & 50.7 & 57.1 & 52.8 & 45.0 & 52.3 & 52.9 & 70.2 & 48.0 & 48.7 & 51.8 & 71.9 & 55.5 & 65.1 & 51.3 & 57.9\\
QAD \shortcite{le2023quality}   & 72.3 & 55.2 & 80.0 & 61.5 & 78.3 & 65.5 & 54.5 & 76.5 & 79.2 & 86.4 & 56.4 & 58.0 & 57.4 & 82.6 & 77.8 & 63.5 & 56.5 & 68.3\\
ODDN (\textbf{ours})                      & 72.1 & 44.1 & 76.8 & 68.1 & 76.5 & 73.3 & 58.0 & 75.6 & 83.5 & 90.8 & 61.1 & 65.9 & 63.9 & 83.5 & 77.0 & 72.9 & 55.0 & \textbf{70.7}\\
\bottomrule
    \end{tabular}
}
  \caption{The \textbf{quality-agnostic} experimental results across 17 well-known datasets under the \textbf{2-class training data setting}.}
  \label{2class Quality-agnostic}
  \vspace{-0.05in}
\end{table*}

\begin{table*}[!ht]
\Huge
    \centering
\setlength{\tabcolsep}{1.5pt}
\resizebox{\textwidth}{18mm}{
%\resizebox{2.1\columnwidth}{!}{
    \begin{tabular}{ c c c c c c c c c c c c c c c c c c  >{\columncolor{lightgray}}c}
     \bottomrule\hline
      %\multirow{3}*{Method} &\multicolumn{18}{c}{ Test Models}\\ 
\multirow{2}*{Method}  & 
\multirow{2}*{ \makecell[c]{ InfoMax- \\  GAN} }  & 
\multirow{2}*{ \makecell[c]{ BE- \\  GAN} }  & 
\multirow{2}*{ \makecell[c]{ Cramer- \\ GAN} }  & 
\multirow{2}*{ \makecell[c]{ Att- \\ GAN } } & 
\multirow{2}*{ \makecell[c]{ MMD- \\  GAN} }  & 
\multirow{2}*{ \makecell[c]{ Rel- \\  GAN} } & 
\multirow{2}*{ \makecell[c]{ S3- \\  GAN} }  & 
\multirow{2}*{ \makecell[c]{ SNG- \\  GAN} }  & 
\multirow{2}*{ \makecell[c]{ STG- \\  GAN } } & 
\multirow{2}*{ \makecell[c]{ Pro- \\  GAN} } &
\multirow{2}*{ \makecell[c]{ Style- \\  GAN} }&
\multirow{2}*{ \makecell[c]{ Style- \\  GAN2} } &
\multirow{2}*{ \makecell[c]{ Big- \\  GAN} } &
\multirow{2}*{ \makecell[c]{ Cycle- \\  GAN} } &
\multirow{2}*{ \makecell[c]{ Star- \\  GAN} } &
\multirow{2}*{ \makecell[c]{ Gau- \\  GAN} } &
\multirow{2}*{ \makecell[c]{ Deep- \\  fake} } &
Mean \\ &&&&&&&&&&&&&&&&&& Acc \\
\bottomrule\hline
MesoNet\shortcite{afchar2018mesonet}    & 58.7 & 45.4 & 63.5 & 62.9 & 62.0 & 50.2 & 48.7 & 58.4 & 64.1 & 55.4 & 52.0 & 48.1 & 53.7 & 63.2 & 62.0 & 49.6 & 51.8 & 54.3\\
FF++\shortcite{rossler2019faceforensics++}      & 68.9 & 29.9 & 82.0 & 63.3 & 80.4 & 67.2 & 55.5 & 75.4 & 82.0 & 93.0 & 61.1 & 59.8 & 57.9 & 80.1 & 78.6 & 67.3 & 51.9 & 67.9\\
F3Net\shortcite{qian2020thinking}      & 62.0 & 43.4 & 65.8 & 53.2 & 64.1 & 56.7 & 55.4 & 58.8 & 67.7 & 92.5 & 76.6 & 62.3 & 56.8 & 60.5 & 71.0 & 71.3 & 51.1 & 63.4\\
MAT\shortcite{zhao2021multi}       & 52.2 & 49.3 & 62.5 & 50.6 & 60.3 & 51.7 & 53.3 & 53.9 & 58.6 & 92.2 & 54.4 & 54.9 & 54.0 & 76.5 & 59.4 & 68.4 & 51.0 & 59.4\\
SBIs \shortcite{shiohara2022detecting}   & 61.3 & 57.4 & 74.8 & 50.3 & 67.5 & 54.6 & 61.5 & 53.2 & 57.1 & 95.9 & 57.2 & 52.9 & 55.4 & 78.3 & 59.3 & 74.6 & 50.7 & 62.6\\
ADD \shortcite{woo2022add}    & 51.0 & 50.2 & 54.4 & 50.3 & 53.4 & 50.7 & 46.2 & 50.5 & 50.9 & 75.8 & 51.4 & 51.6 & 52.7 & 72.6 & 52.3 & 66.4 & 50.7 & 55.0\\
QAD \shortcite{le2023quality}     & 76.7 & 46.4 & 79.6 & 68.5 & 77.1 & 73.6 & 58.3 & 76.3 & 81.0 & 90.2 & 65.3 & 71.3 & 64.6 & 81.8 & 77.1 & 66.7 & 55.1 & 71.0\\
ODDN (\textbf{ours})                    & 80.4 & 35.1 & 81.0 & 68.7 & 78.2 & 74.5 & 62.2 & 77.5 & 81.7  & 91.7 & 69.2 & 70.4 & 68.0 & 78.8 & 73.4 & 73.8 & 55.3 & \textbf{72.1}\\
\bottomrule
    \end{tabular}
}
  \caption{The \textbf{quality-agnostic} experimental results across 17 well-known datasets under the \textbf{4-class training data setting}.}
  \label{4class Quality-agnostic}
  \vspace{-0.18in}
\end{table*}

\section{Experiments}
In this section, we provide an overview of the comprehensive experiments designed to thoroughly evaluate the effectiveness of the proposed ODDN. We first introduce the experimental settings, including the dataset and implementation details. Then, we provide a detailed account of the preliminary experiment and various grouping experiments.
\subsection{Settings}
\textbf{Datasets:} To ensure a consistent basis for comparison, we employ the training set from \textit{ForenSynths} to train the detectors, in line with the baselines \cite{wang2020cnn,jeong2022bihpf,jeong2022frepgan}. This training set comprises 20 distinct categories, each featuring 18,000 synthetic images generated using ProGAN, alongside an equal number of real images sourced from the LSUN dataset. For evaluation, we utilized a comprehensive collection of 17 commonly used datasets. The first 8 datasets are derived from the \textit{ForenSynths} \cite{wang2020cnn}, including images generated by eight distinct generation models. The remaining 9 datasets are derived from the \textit{GANGen-Detection} \cite{chuangchuangtan-GANGen-Detection}, comprising images generated by nine additional GANs.

\noindent\textbf{Implementation Details:}
We use the Adam \cite{kingma2015adam} as the optimizer with a learning rate of \(2 \times 10^{-4}\) and a batch size of 128. For the hyper-parameter $\alpha$, we adhere to the traditional setting, namely 0.004. In our framework,  encoder can be any standard image classifier, such as Res50, to extract features from the image. In order to maintain consistency with previous endeavors\cite{le2023quality}, we employ ResNet-50 (Res50) as our encoder. Our method is implemented using PyTorch on Nvidia GeForce RTX 3090 GPU. We adhere to the commonly used evaluation metrics accuracy (Acc), following common researches.

\subsection{Quality-aware Experiments}
Following the classic setting, we utilize two groups of training sets: a 2-class set (``chair" and ``horse") and a 4-class set (``car", ``cat", ``chair", and ``horse") from the \textit{ForenSynths} dataset. The results are presented in Table \ref{2class Quality-aware} and \ref{4class Quality-aware}. To emulate the composition of OSN data in open scenarios, 20\% of the data were compressed using operations adopted by popular OSN with constant rate quantization parameters of 40 to create paired data. The remaining 80\% were unprocessed to simulate scenarios where unpaired data is significantly more prevalent than paired data in OSN. It should be clarified that if there are baselines specifically designed for paired data, unpaired data should also be utilized for classification purposes, rather than being left idle, to ensure a fair comparison. During the inference stage, we compress entire images of the testing set by the same compression as the training set and subsequently evaluate each compressed image.

The 2-class and 4-class experimental results presented in Table \ref{2class Quality-aware} and \ref{4class Quality-aware} compare the performance of various detection methods across 17 different datasets, using the accuracy metric (Acc) as the primary evaluation metric. As shown in Table \ref{2class Quality-aware}, in the 2-class experiment, the proposed ODDN achieved the highest mean accuracy of 71.4\%, significantly outperforming other methods. For instance, QAD, the second-best performer, achieved a mean accuracy of 69.2\%, while FF++ and F3Net had mean accuracies of 67.3\% and 63.2\%, respectively. The proposed method showed particularly strong performance with specific GANs such as ProGAN, STGGAN, and CycleGAN, achieving accuracies of 91.3\%, 85.0\%, and 80.8\%, respectively. This indicates that the method is highly effective in distinguishing between real and fake images in a binary classification setup. Moreover, the method consistently performed well across most datasets, achieving over 70\% accuracy in 10 out of the 17. This consistency across various datasets shows the robustness and reliability of the ODDN in quality-aware scenarios.

As shown in Table \ref{4class Quality-aware}, In the 4-class quality-aware experiment, the proposed method again demonstrated superior performance with the highest mean accuracy of 72.6\%, further confirming its effectiveness in more complex classification tasks. QAD followed closely with a mean accuracy of 70.9\%, maintaining its position as a strong competitor. Other methods like FF++ and F3Net achieved mean accuracies of 68.7\% and 62.4\%, respectively, indicating a noticeable performance gap between these methods and the top performers. The proposed method excelled in detecting images from GANs like ProGAN, STGGAN, and CycleGAN, with accuracies of 94.0\%, 85.8\%, and 84.9\%, respectively. These high accuracies highlight the deepfake detection capability of the proposed DANN in another training data scenario. Additionally, DANN showed consistent high performance across most datasets, achieving over 70\% accuracy in 11 out of the 17 well-known datasets. This consistency and robustness make it a reliable choice for quality-aware analysis of the generated images in binary classification tasks.

Overall, the experimental results indicate that the proposed method is highly effective and reliable for detecting GAN-generated images. Its superior performance and consistency across different GANs and classification tasks make it a standout choice for quality-aware analysis. The method's ability to achieve high accuracies in both 2-class and 4-class experiments suggests that it can effectively distinguish between real and fake images, regardless of the complexity of the task. This robustness highlights the potential for practical applications where detecting GAN-generated content.

\begin{table*}[!t]
\vspace{-0.4in}
\Huge
\centering
\setlength{\tabcolsep}{3pt}
\resizebox{\textwidth}{9.2mm}{
\begin{tabular}{ c c c c c c c c c c c c c c c c c c  >{\columncolor{lightgray}}c}
\bottomrule\hline
\multirow{2}*{Method}  & 
\multirow{2}*{ \makecell[c]{ InfoMax- \\  GAN} }  & 
\multirow{2}*{ \makecell[c]{ BE- \\  GAN} }  & 
\multirow{2}*{ \makecell[c]{ Cramer- \\ GAN} }  & 
\multirow{2}*{ \makecell[c]{ Att- \\ GAN } } & 
\multirow{2}*{ \makecell[c]{ MMD- \\  GAN} }  & 
\multirow{2}*{ \makecell[c]{ Rel- \\  GAN} } & 
\multirow{2}*{ \makecell[c]{ S3- \\  GAN} }  & 
\multirow{2}*{ \makecell[c]{ SNG- \\  GAN} }  & 
\multirow{2}*{ \makecell[c]{ STG- \\  GAN } } & 
\multirow{2}*{ \makecell[c]{ Pro- \\  GAN} } &
\multirow{2}*{ \makecell[c]{ Style- \\  GAN} }&
\multirow{2}*{ \makecell[c]{ Style- \\  GAN2} } &
\multirow{2}*{ \makecell[c]{ Big- \\  GAN} } &
\multirow{2}*{ \makecell[c]{ Cycle- \\  GAN} } &
\multirow{2}*{ \makecell[c]{ Star- \\  GAN} } &
\multirow{2}*{ \makecell[c]{ Gau- \\  GAN} } &
\multirow{2}*{ \makecell[c]{ Deep- \\  fake} } &
Mean \\ &&&&&&&&&&&&&&&&&& Acc \\
\bottomrule\hline

Baseline                     & 74.1 & 40.9 & 78.9 & 66.9 & 77.0 & 70.6 & 57.4 & 77.3  & 84.8 & 90.8 & 58.8 & 61.4 & 60.1 & 81.8 & 79.7 & 70.6 & 53.1 & 69.4 \\
+ ODA                     & 75.2 & 39.0 & 80.2 & 63.9 & 78.9 & 73.3 & 59.7 & 77.2 & 84.3 & 92.8 & 62.8 & 62.7 & 62.3 & 83.3 & 80.0 & 78.1 & 54.1 & 71.0\\
+ ODA + CGC (\textbf{ours})                  & 73.1 & 42.3 & 76.1 & 71.2 & 75.9 & 72.5 & 60.5 & 75.5 & 85.0 & 91.3 & 64.5 & 69.4 & 64.3 & 80.8 & 78.0 & 77.3 & 54.3 & \textbf{71.4}\\
\bottomrule
\end{tabular}}
\caption{The experimental results of the ablation study. The settings are the same as 2-class quality-aware experiment above.}
\label{Ablation:ACC}
\vspace{-0.18in}
\end{table*} 
\subsection{Quality-agnostic Experiments}
In this group of quality-agnostic experiments, the training settings are the same with the above quality-aware experiments, that is 2-class and 4-class. It should be noted that the test images used here do not follow the compression applied to the training data. Instead, they are compressed using JPEG compression coefficients sampled from a normal distribution ranging from 30 to 100, simulating open-world scenarios that need to handling unknown compression methods. The evaluation results of the 2-class and 4-class quality-agnostic experiments are shown in Table \ref{2class Quality-agnostic} and \ref{4class Quality-agnostic}.

In the 2-class quality-agnostic experiment, the proposed ODDN demonstrated remarkable performance with the highest mean accuracy of 70.7\%. This indicates its robustness in identifying real versus fake images without accounting for the quality of the generated images. Specifically, it excelled with GANs such as ProGAN (90.8\% accuracy), STGGAN (83.5\%), and CycleGAN (83.5\%). This high level of performance across diverse GANs underscores the method's adaptability and effectiveness. QAD was the second-best performer with a mean accuracy of 68.3\%, making it a reliable alternative but still falling short of the proposed method's overall effectiveness. Other methods like FF++ and F3Net had mean accuracies of 66.0\% and 59.3\%, respectively, indicating a significant performance gap between these methods and the top performers. These results suggest that while multiple methods are viable for quality-agnostic GAN detection, the proposed ODDN stands out for its consistent and superior performance across baselines.

In the 4-class quality-agnostic experiment, the proposed ODDN again outperformed others with the highest mean accuracy of 72.1\%. It achieved notable accuracies with GANs such as ProGAN (91.7\%), STGGAN (81.7\%), and CycleGAN (78.8\%), further demonstrating its robustness in handling more complex classification tasks. QAD followed closely with a mean accuracy of 71.0\%, reinforcing its reliability but still trailing behind the proposed method. FF++ and F3Net had mean accuracies of 67.9\% and 63.4\%, respectively, which, while respectable, highlight the superior consistency and accuracy of the proposed method.

\begin{figure}[h]
\centering
\vspace{0.1in}
\includegraphics[width=1\columnwidth]{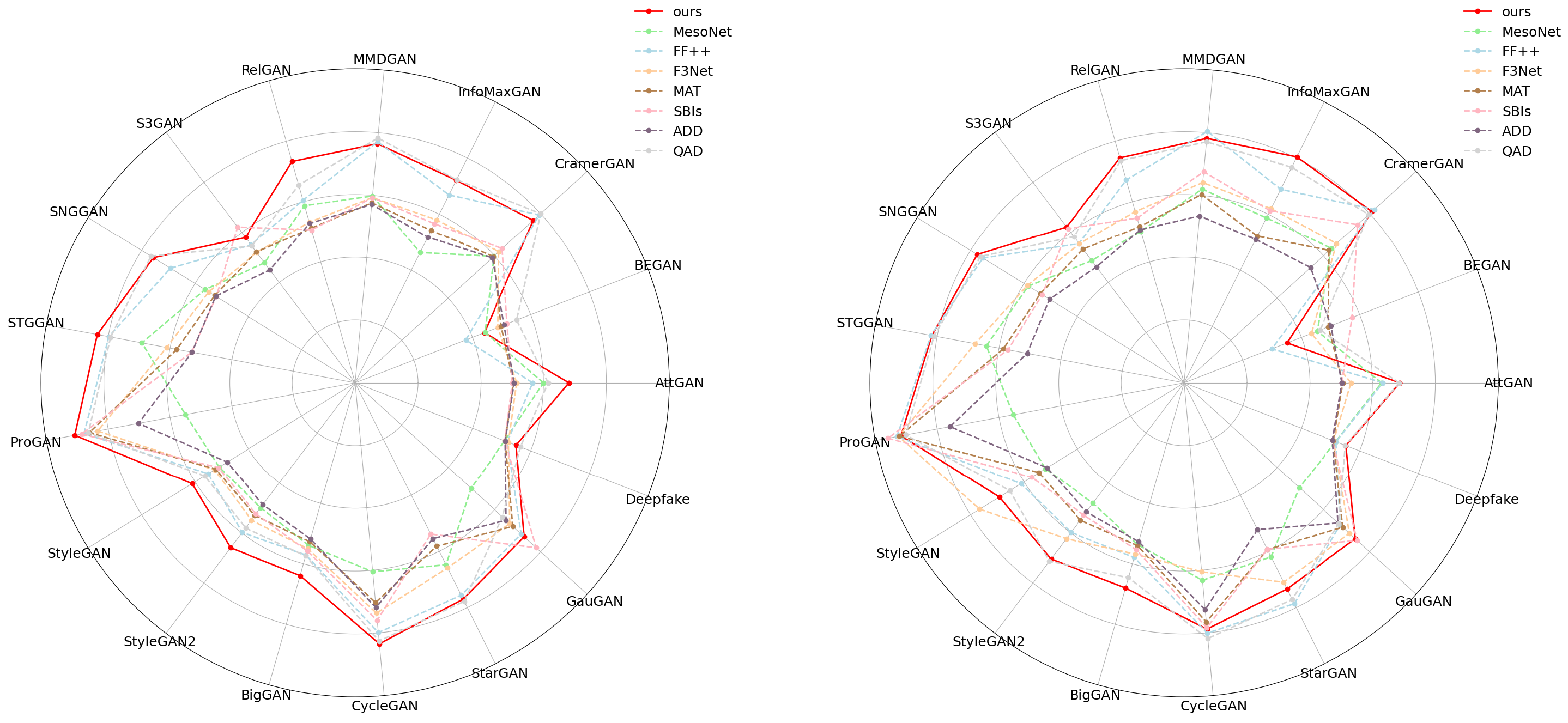}
\caption{Performance comparison across 17 well-known datasets in quality-agnostic experiments (simulating the open-world OSN scenario) is illustrated for both the 2-class (left figure) and 4-class (right figure) training settings. The ODDN we proposed outperforms these baselines overall.}
\label{fig-leida}
\vspace{-0.1in}
\end{figure}

As illustrated in Figure \ref{fig-leida}, these results underscore the ODDN's ability to deliver high performance consistently, making it a highly effective choice for quality-agnostic analysis of deepfakes. The robustness across 17 different well-known datasets in different training settings suggests its potential for practical applications in open-world scenarios where distinguishing GAN-generated content is crucial.

\subsection{Ablation Study}
The ablation study presented in the table evaluates the impact of different components (ODA and CGC) on the performance of the baseline method across the 17 datasets, using mean accuracy (Acc) as the metric. The results are analyzed in three configurations: the baseline, the baseline with ODA, and the baseline with both ODA and CGC.

The baseline achieved a mean accuracy of 69.4\%, showing strong performance across several GANs. Notably, it performed exceptionally well with ProGAN (90.8\%), STGGAN (84.8\%), and CycleGAN (81.8\%). However, there were GANs where the baseline method's performance was less impressive, such as AttGAN (66.9\%) and BEGAN (40.9\%). Adding ODA to the baseline resulted in an improvement in mean accuracy, increasing to 71.0\%. This enhancement indicates that ODA positively contributes to the model's ability to distinguish between real and fake images. Specific datasets like InfoMaxGAN, MMDGAN, and StyleGAN2 saw noticeable improvements, with accuracy increasing to 75.2\%, 78.9\%, and 62.8\%, respectively. The improvements were consistent across most datasets, demonstrating the robustness of the ODA component. Further adding CGC to the baseline with ODA configuration led to the highest mean accuracy of 71.4\%. This configuration achieved the best performance across datasets, indicating that the combination of ODA and CGC significantly enhances the model's overall accuracy. The method excelled particularly with ProGAN (91.3\%), STGGAN (85.0\%), and CycleGAN (80.8\%). The combined approach also improved performance in datasets where the baseline method had lower accuracy, such as BE-GAN and Cramer-GAN, demonstrating its effectiveness in a broader range of scenarios.

The ablation study clearly shows that both the proposed ODA and CGC modules contribute positively to the model's performance. This comprehensive analysis highlights the effectiveness of integrating ODA and CGC into the baseline for achieving superior performance across various datasets.

\subsection{Feature Distribution Visualization}
\begin{figure}[h]
\centering
\includegraphics[width=1\columnwidth]{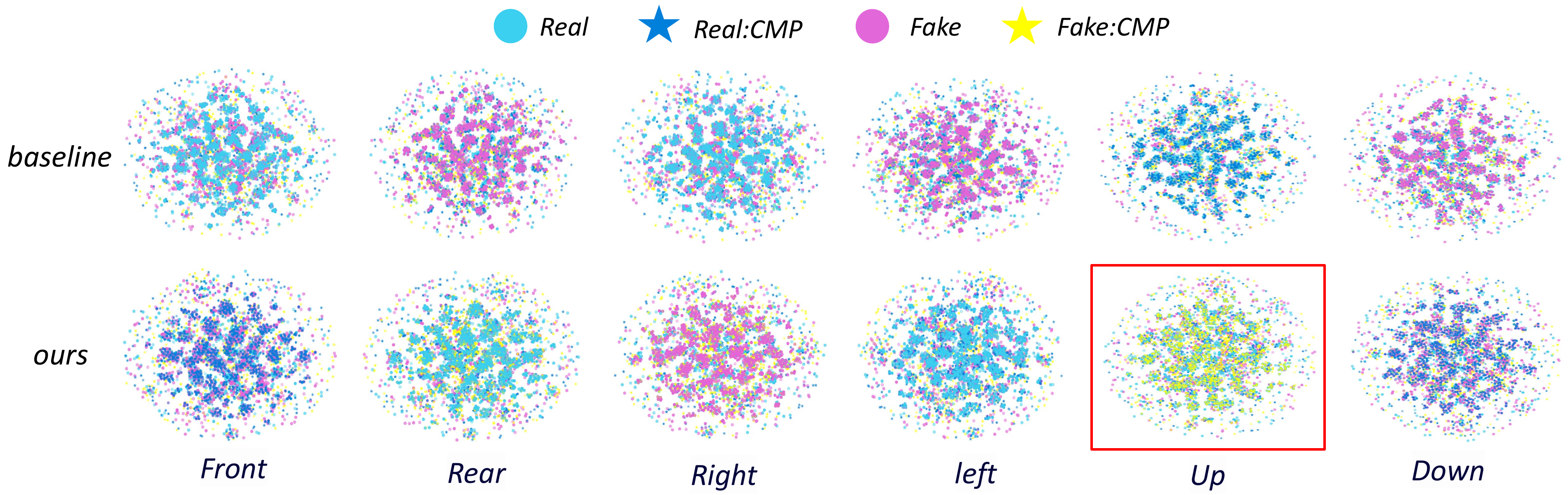} % Reduce the figure size so that it is slightly narrower than the column. Don't use precise values for figure width.This setup will avoid overfull boxes.
\caption{The feature visualization of baseline and ODDN.}
\label{fig-vis}
\vspace{-0.18in}
\end{figure}

To verify the consistency of invariant representation across input quality, we visualized the feature distribution of the baseline and ODDN using t-SNE \cite{van2008visualizing} in 3D, observing from six angles: front, rear, right, left, up, and down (Figure \ref{fig-vis}). For the baseline model, compressed deepfake features tend to cluster closely with other features, making them difficult to distinguish from multiple perspectives. This close proximity is likely a key reason for the baseline's reduced detection performance. In contrast, ODDN significantly increases the separation between features of different classes, with each class clearly occupying distinct regions in at least one of the six observed directions. This greater separation allows for more effective distinction between features, leading to improved detection accuracy. In summary, the proposed ODDN demonstrates superior generalization across varying input qualities, enhancing its performance in distinguishing deepfakes.

\section{Conclusion}
In conclusion, this paper presents the Open-world Deepfake Detection Network (ODDN), a novel approach designed to address the challenges of deepfake detection in open-world scenarios, particularly on online social networks where unpaired data is prevalent. Through the integration of two key modules: open-world data aggregation
(ODA) and compression-discard gradient correction
(CGC), ODDN effectively handles the complexities associated with varying data qualities and compression methods. The comprehensive experiments are conducted across 17 popular datasets under diverse test settings demonstrate that ODDN significantly outperforms existing SOTA models, showcasing its robustness and adaptability in real-world applications. This work not only advances the field of deepfake detection but also provides a valuable benchmark for future researches aimed at combating misinformation on the online social platforms.

\bibliography{aaai25}
\end{document}